\documentclass[final, 3p, times]{elsarticle}

\usepackage[utf8]{inputenc} 
\usepackage[T1]{fontenc}    
\usepackage{url}            
\usepackage{booktabs}       
\usepackage{amsfonts}       
\usepackage{nicefrac}       
\usepackage{microtype}      
\usepackage{soul}
\usepackage{color}
\usepackage{amsmath}
\usepackage{graphicx}
\usepackage{subfigure}
\usepackage{algorithm}
\usepackage{algorithmic}

\begin{document}

\begin{frontmatter}

\title{Low-Cost Recurrent Neural Network Expected Performance Evaluation}

\author[uma]{Andr\'{e}s Camero\corref{cor} }
\ead{andrescamero@uma.es}
\ead[url]{http://neo.lcc.uma.es/staff/acamero/} 
\cortext[cor]{Corresponding author.}
\author[mit]{Jamal Toutouh}
\ead{toutouh@mit.edu}
\author[uma]{Enrique Alba}
\ead{eat@lcc.uma.es}
\address[uma]{Departamento de Lenguajes y Ciencias de la Computaci\'{o}n, Universidad de M\'{a}laga, Andaluc\'{i}a Tech, Espa\~{n}a}
\address[mit]{Massachusetts Institute of Technology, Computer Science \& Artificial Intelligence Laboratory, Cambridge, MA, USA}

\begin{abstract}
Recurrent neural networks are a powerful tool, but they are very sensitive to their hyper-parameter configuration. 
Moreover, training properly a recurrent neural network is a tough task, therefore selecting an appropriate configuration is critical. 
Varied strategies have been proposed to tackle this issue. However, most of them are still impractical because of the time/resources needed. 
In this study, we propose a low computational cost model to evaluate the expected performance of a given architecture based on the distribution of the error of random samples of the weights. 
We empirically validate our proposal using three use cases. The results suggest that this is a promising alternative to reduce the cost of exploration for hyper-parameter optimization.
\end{abstract}

\begin{keyword}
deep learning \sep recurrent neural network \sep hyper-parameter evaluation
\end{keyword}

\end{frontmatter}

\section{Introduction}

One of the most promising trends in Machine Learning (ML) focuses on learning features from data through multiple layers of abstraction. This trend is known as Deep Learning (DL)~\cite{Lecun2015}. The main idea behind this new sub-field of Artificial Intelligence (AI) is that Deep Neural Networks (DNNs) learn hierarchical layers of representation when applying them together with sufficient computing resources and data.

In this study, we focus on Recurrent Neural Networks (RNNs), a special case of DNNs. Their architecture  consists of both feedforward and feedback connections between layers and neurons~\cite{Lipton2015}. This feature enables RNNs to capture long-term dependencies in an input sequence. Thus, they have exceeded their competitors in addressing ML tasks that involve sequential modeling and prediction, such as handwriting recognition or language modeling~\cite{Lecun2015}. However, like the rest of DNNs, RNNs present a big drawback: the difficulty of their learning process. Particularly, the RNN architecture has two well-known issues: the \emph{vanishing} and the \emph{exploding} gradient~\cite{bengio1994learning,Pascanu2013}. 

In order to improve the DNNs learning process, a promising line of research bears on  optimizing (modifying and adapting) the hyper-parameters that define the neural network  architecture to the specific datasets by changing the activation functions, increasing or reducing the number of hidden layers, and modifying the kernel size of a layer, among others~\cite{Bergstra2011,Camero2018,Jozefowicz2015}. The most widely applied methods for hyper-parameter selection in DL are the manual exploration by human experts and the automatic search by using specific algorithms (e.g., grid or random search). 

The automatic hyper-parameter selection methods are principally based on \textit{trial-and-error} through a high-dimensional hyper-parameter space. Therefore, they are computationally inefficient due to the high cost of the DL algorithms used to evaluate every tentative hyper-parameter configuration. For this reason, it is necessary to speed up the evaluation of the architecture configuration to improve the competitiveness of the automatic hyper-parameter selection algorithms~\cite{Domhan2015}.

In this study, we propose to create a method to evaluate the performance of a given network architecture inspired by the linear time-invariant theory~\cite{allen2004signal}. The idea is to study the output of a network by randomly generating sets of weights, and using the results to infer the performance of the network without actually training the RNN. The main advantages of this approach are: (\textit{a}) it requires critically less computational resources, and therefore, it can be exhaustively used by an automatic \mbox{hyper-parameter} selection method to evaluate a high number of architectures; and (\textit{b}) it can be easily generalized to evaluate the architecture of any other DNN. 

Therefore, the main contribution of this study is to define a general low computational cost method to evaluate the performance of a DNN hyper-parameter configuration (architecture), which speeds up any automatic hyper-parameter search approach. We test our proposal using three datasets:
a sine wave, the occupancy rate of the car parks in a real city, and the appliances energy consumption of a house (real data). And finally, we discuss the validity of the results.

The remainder of this paper is organized as follows: the next section briefly reviews some of the most relevant works related to our proposal. Section~\ref{section:sampling} introduces the error sampling method to evaluate RNN architectures. Section~\ref{section:experiments} presents the experimental analysis of our proposal over the three use cases. Section~\ref{section:discussion} discusses the main advantages and drawbacks of our proposal. 
Finally, Section~\ref{section:conclusions} outlines the conclusions and proposes future work.

\section{Related Work}\label{section:related}

RNNs are feedforward neural networks extended by incorporating edges that span adjacent time steps, introducing a notion of time to the model. These special type of edges, known as recurrent edges, may form cycles and connect a node to itself (self-connections) across time. Thus, at a time $t$, a node connected to recurrent edges receives input from the current data point $x^{t}$ also from hidden node $h^{t-1}$ (the previous state of the network). The output $y^{t}$ at each time $t$ is computed according to the hidden node values at time $t$ ($h^t$). Input at time $t-1$ ($x^{t-1}$) can determine the output at time $t$ ($y^{t}$) by way of recurrent connections~\cite{Lipton2015}.  

In general, RNNs are difficult to train by gradient-based optimization procedures, e.g., stochastic gradient descent or second-order methods, due to the exploding and the vanishing gradient problems~\cite{bengio1994learning,Kolen2001,Pascanu2013}.  

Recently, learning systems based on Long Short-Term Memory (LSTM)~\cite{Hochreiter1997} have demonstrated to be resistant to gradient problems when dealing with learning tasks as image captioning, handwriting recognition or language translation.  Furthermore, recent studies applied optimization techniques to enable large-scale learning in specific applications of RNNs~\cite{Bengio2013}, since they have shown competitive results in another type of neural networks.

Weight initialization determines the learning rate, the convergence rate, and the probability of classification error. Thus, Ramos et al.~\cite{Ramos2017} developed a theory to objectively test different weight initialization strategies and provide a quantitative measure for each weight initialization strategy.  

In the last recent years, selecting appropriate hyper-parameters for the neural network to a given dataset, instead of using a generalized one, has been analyzed to improve the learning process, showing competitive results~\cite{Bergstra2011,Jozefowicz2015}. These hyper-parameters define the number of layers, the number of hidden unit per layer, the activation function, the kernel size of a layer, etc. 

Intelligent automatic hyper-parameter methods are showing to be competitive in choosing neural network settings in comparison to those provided by human experts. However, these methods are still not generally adopted since they are not efficient in searching in a high-dimensional hyper-parameter space (they require high computational resources). Mainly because (in general) these methods are data driven, and therefore, they require fitting a model and evaluating its performance on validation data, which can be an expensive process~\cite{Albelwi2017,Bergstra2011,Smithson2016}. 

Few authors have proposed strategies to address this issue. Domhan et al.~\cite{Domhan2015} presented a method for speeding up the hyper-parameter search by detecting and finishing evaluations that under-perform previously computed hyper-parameterizations. Even though they reduced the hyper-parameterization optimization process time down to 50\%, the required computation times were still too long to be practical. 

In contrast to the last approach, we propose and analyze a method to evaluate the hyper-parameters of a given RNN architecture avoiding the high-cost learning (weights optimization) process. Instead of applying a gradient-based optimization procedure to compute the weights, the proposed method uses an \textit{error sampling} approach, i.e., it generates a representative sample of random sets of weights, which are used to compute the error of a given architecture on a test set. This error sampling allow us to reduce the computational cost, cutting down the time of the automatic hyper-parameter methods.

\section{Ranking Based on the Error Sampling}\label{section:sampling}

Given an RNN architecture, and a time series, it is easy to show that the output of the network changes as the weights change~\cite{haykin2009neural}. Therefore, we may ask ourselves: is it possible to characterize the \emph{response} of this architecture to the input? A naive approach to this characterization is to study the variation of the observed output in regards to the expected one (i.e., the error) as the weights change, an idea inspired in signal analysis~\cite{allen2004signal}. Particularly, we assume that the error of each architecture may be described by a probability density function (PDF). Therefore, we can estimate the probability $p_t$ of finding a set of weights whose error is below a defined \emph{threshold}; then as this probability increases, it would be likely to observe a better result after training the network. If this were possible, given a set of architectures, we propose to rank them by their probability $p_t$.

\begin{figure}[!h]
  \centering
  \includegraphics[width=0.9\textwidth]{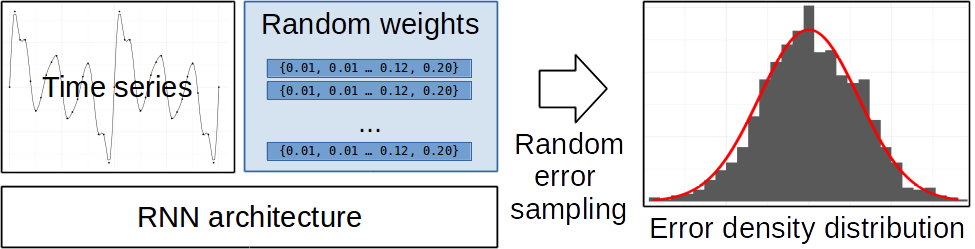}
  \vspace{-0.3cm}
  \caption{Error sampling based on random weight initialization}
  \label{figure:mae-sampling}
\end{figure}

First, we take multiple samples of the output of a specific RNN architecture, where each sample is obtained using a random set of weights. Then, for each sample, we compute the error. Finally, we approximate the density of the errors sampled with a (known) PDF and estimate $p_t$. Note that we are \emph{preprocessing} the hyper-parameter space, avoiding the training of all configurations, in favor of selecting a few candidates (to be trained in the future). Figure~\ref{figure:mae-sampling} presents a high-level view of our proposal: the \emph{random error sampling}.

Specifically in this study, as a first approach to explore our proposal, we propose to perform the \emph{random error sampling} using the mean absolute error (MAE) measure~\cite{hyndman2006another}, initializing the weights using a $\mathcal{N}(0,1)$ distribution~\cite{Ramos2017}, and approximating the errors density with a truncated normal distribution. We will refer to this combination as the \emph{MAE random sampling}.


Algorithm~\ref{algorithm:sampling} presents the pseudo-code of the MAE random sampling. Given a number of LSTM cells (\textit{NC}), a number of time steps or look back (\textit{LB}), and a user-defined time series (data), the algorithm takes \textit{MAX\_SAMPLES} samples of the MAE with normally distributed weights. Once the sampling is done, we fit a truncated normal distribution to the MAE values sampled, and finally, we estimate $p_t$ for the defined \textit{THRESHOLD}. To rank a set of architectures we have to compute $p_t$ for each of them and use it to sort them.

\begin{algorithm}[!h]
	\caption{MAE random sampling pseudo-code.}
	\label{algorithm:sampling}
	\begin{algorithmic}[1]
		\STATE {\textit{data} $\gets$ LoadData()}
        \STATE {\textit{rnn} $\gets$ InitializeRNN(\textit{NC}, \textit{LB})}
        \STATE {\textit{mae} $\gets \emptyset$ }
        \WHILE {sample $\leq$ \textit{MAX\_SAMPLES} } 
            \STATE {\textit{weights} $ \gets $ GenerateNormalWeights(0,1)}
            \STATE {UpdateWeights(\textit{rnn}, \textit{weights})}
            \STATE {\textit{mae}[sample] $\gets$ MAE(\textit{rnn}, \textit{data})}	
            \STATE{ sample++ }
		\ENDWHILE
        \STATE {\textit{mean, sd} $\gets$ FitTruncatedNormal(\textit{mae}) }
        \STATE {$p_t \gets$ PTruncatedNormal(\textit{mean}, \textit{sd}, \textit{THRESHOLD}) }
	\end{algorithmic}
\end{algorithm}

\section{Experiments}\label{section:experiments}

We have implemented our proposal\footnote{Code available in Github} 
in Python, using DLOPT~\cite{camero2018dlopt}, Keras~\cite{chollet2015keras} and Tensorflow~\cite{abadi2016tensorflow}. Then, we studied three use cases: (\emph{i}) a sine wave, (\emph{ii}) the occupancy rate of 29 car parks located in Birmingham, UK, and (\emph{iii}) the appliances energy consumption of a house in Stambruges, Belgium, and the nearby weather data~\cite{candanedo2017data}. In all cases we sampled one hidden layer stacked RNN architectures~\cite{haykin2009neural}, using the parameters defined in Table~\ref{table:parameters}.

\begin{table}[!h]	
    \caption{MAE sampling parameters}    
    \label{table:parameters}
    \centering    
    \begin{tabular}{ rrrr }
    \toprule
    NC & LB & MAX\_SAMPLES & THRESHOLD \\
    \midrule
    100 & 30 & 1000 & 0.01 \\
    \bottomrule
    \end{tabular}
\end{table}

\subsection{Sine Wave}

A sine wave is a mathematical curve that describes a smooth periodic oscillation. Despite its simplicity, we are interested in its study, because adding sine waves it is possible to approximate any periodic waveform~\cite{bracewell1986fourier}. Equation~\ref{equation:sine} expresses a sine wave as a function of time (\emph{t}), where $A$ is the peak amplitude, $f$ is the frequency, and $\phi$ is the phase.

\vspace{-0.3cm}
\begin{align}\label{equation:sine}
y(t) = A \cdot sin(2 \pi \cdot f \cdot t + \phi)
\end{align}

\begin{figure}[!h]
  \centering
  \subfigure[Frequency=1]{
  	\label{figure:sine-wave-f1}
  	\includegraphics[width=0.45\textwidth]{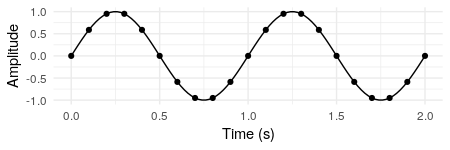}}
  \subfigure[Frequency=3]{
  	\label{figure:sine-wave-f3}
    \includegraphics[width=0.45\textwidth]{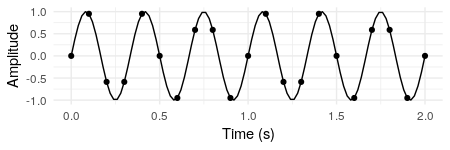}}
  \vspace{-0.3cm}
  \caption{Sine wave sampling (10 samples per second)}
  \label{figure:sine-waves}
\end{figure}

We sampled a sine wave using the following parameters: $A=1$, $f=1$, and $\phi=0$, in the range $t \in [0,100]$~seconds (s), and at 10 samples per second. Figure~\ref{figure:sine-wave-f1} shows the continuous wave along with the sampled wave (dots). 

\begin{figure}[!h]
  \centering
  \includegraphics[width=0.9\textwidth]{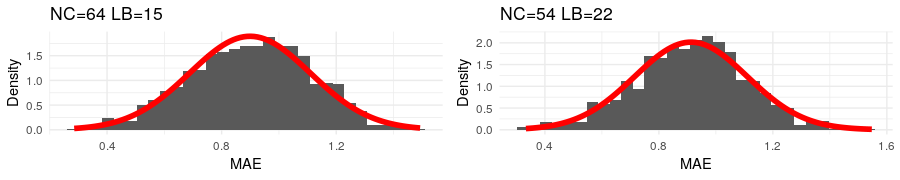}
  \caption{MAE samples density distribution and its truncated normal approximation}
  \label{figure:sine-f1-trunc-norm}
\end{figure}

Then, for each architecture (\textit{NC} and \textit{LB}) defined in Table~\ref{table:parameters}, we sampled the MAE. Since $y(t) \in [-1,1]$, we set the activation function of the output layer to be a $tanh$, therefore the $MAE \in [0,2]$. Given so, we fitted a truncated normal distribution in the range [0,2] to the samples and calculated the probability $p_{0.01}$. Figure~\ref{figure:sine-f1-trunc-norm} presents two MAE sampling distribution examples. NC stands for the number of LSTM cells in the hidden layer, and LB for the look back parameter of the configuration sampled.

\begin{figure}[!h]
  \centering
  \includegraphics[width=\textwidth]{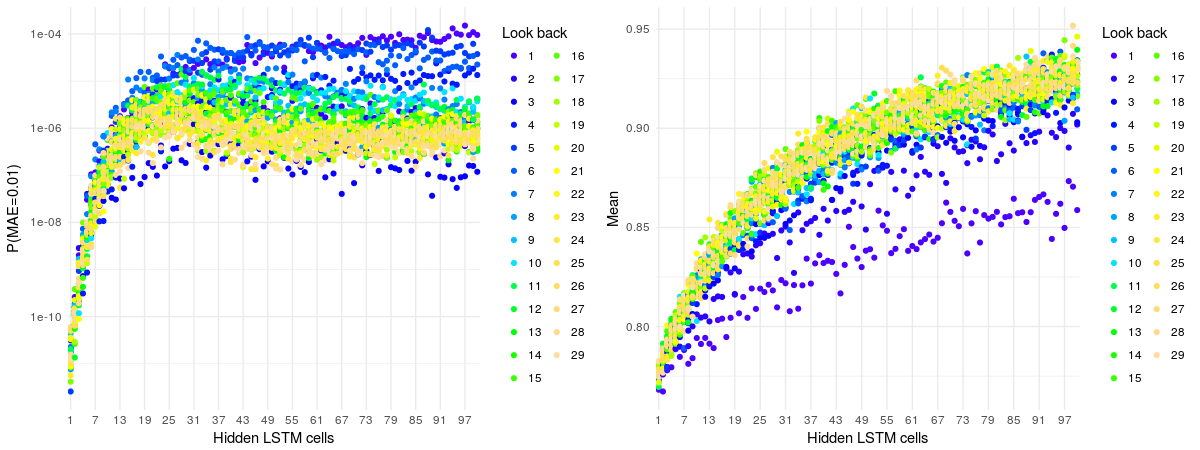}
  \caption{Probability of a configuration whose MAE is equal to 0.01, and the mean of the samples}
  \label{figure:sine-f1-p-mean}
\end{figure}

Figure~\ref{figure:sine-f1-p-mean} summarizes the calculated probabilities $p_{0.01}$ and the mean of the truncated normal distributions fitted. It is quite interesting that $p_{0.01}$ rapidly increases its value until 25 cells are added to the net, afterward, the probability seems to reach a plateau.

To explore the validity of our proposal, we sorted the architectures by $p_{0.01}$ into deciles. Then, we trained ten architectures from each decile (i.e., 100 architectures in total) using the Adam optimizer~\cite{kingma2014adam}
and computed the MAE. Table~\ref{table:sine-f1-bp-corr} presents the correlation between the MAE of the architectures trained using Adam (epochs) and the results of the MAE random sampling (mean, sd, and log $p_{0.01}$). The table also presents the correlation between the training results and the number of LSTM cells (NC) and the timesteps (LB). There is a moderate to strong negative correlation~\cite{rubin2012statistics} between the sampling results and the \emph{training} results. In other words, a good training result (small value) is correlated to a higher value of $p_{0.01}$.

\begin{table}[!h]	
    \caption{Correlation between the MAE of an RNN trained and the random sampling (sine)}    
    \label{table:sine-f1-bp-corr}
    \centering    
    \begin{tabular}{ lrrrrr }
    \toprule
    Epochs & NC & LB & Mean & Sd & log $p_{0.01}$ \\ 
    \midrule
	1 & -0.410 & -0.058 & -0.447 & -0.317 & -0.197 \\ 
    10 & -0.783 & -0.086 & -0.726 & -0.431 & -0.297 \\ 
	100 & -0.561 & -0.274 & -0.790 & -0.641 & -0.638 \\ 
    1000 & -0.454 & -0.164 & -0.668 & -0.458 & -0.502 \\ 
    \bottomrule
    \end{tabular}
\end{table}

The results also suggest that if the training time (\emph{epochs}) is limited (less than 10), the number of hidden neurons tends to be the most important predictor. On the other hand, if we give more time to the training process, then the probability arises as a good predictor. However, it also seems that as we continue increasing the training time, the method loses reliability. This phenomenon might be explained by the method definition itself. Indeed, the MAE random sampling is estimating the probability of a good result (i.e., the architecture suits the problem), but it is not predicting the error after the learning process.

\begin{figure}[!h]
  \centering
  \includegraphics[width=\textwidth]{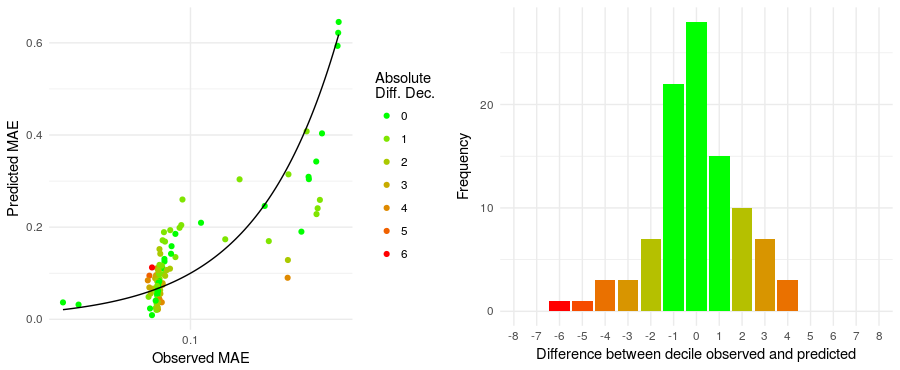}
  \caption{Difference between the predicted and the observed MAE and decile for $f=1$ (sine)}
  \label{figure:sine-f1-pred-obs}
  \vspace{-0.3cm}
\end{figure}

Using 80\% of the architectures (random sampled), we fitted a linear model to the MAE based on the sampling results. Then, we evaluated the accuracy of the model to predict the MAE of the remainder architectures. We measure the difference between the observed and predicted value, and then we ranked the predictions into deciles, and compared the observed and predicted deciles. We repeated this procedure 30 times. On average, the residual standard error of the predicted values was equal to 0.077, while the Spearman correlation coefficient between the predicted and observed decile was equal to 0.784 ($p$-value =0.0006). It is important to notice that the minimum observed MAE (after training for 100 epochs) was equal to 0.021, the maximum 0.612, and the mean 0.133. Overall, on average, 62\% of the fitted values differ at most one decile. Figure~\ref{figure:sine-f1-pred-obs} summarizes the results of a linear model used to predict the MAE. 

To evaluate the generalization capability of the proposal, we repeated the whole process (i.e., MAE random sampling, truncated normal fitting, sorting, and linear fitting) using the sine wave described by $A=1$, $f=3$ and $\phi=0$ (see Figure~\ref{figure:sine-wave-f3}), sampled at 10 samples per second in the range $t \in [0,10]$ (s). Then, we trained a sample of the architectures (100 epochs).
We evaluated 30 linear models, where 80\% of the architectures were used to fit the model, and 20\% to evaluate. On average, the residual standard error of the predicted values was equal to 0.077, while the Spearman correlation coefficient between the predicted and observed decile was equal to 0.663 ($p$-value =0.011). These results confirm the suitability of our proposal for ranking architectures.

\subsection{Car Park Occupancy Rate of Birmingham}

\begin{figure}[!h]
  \centering
  \includegraphics[width=\textwidth]{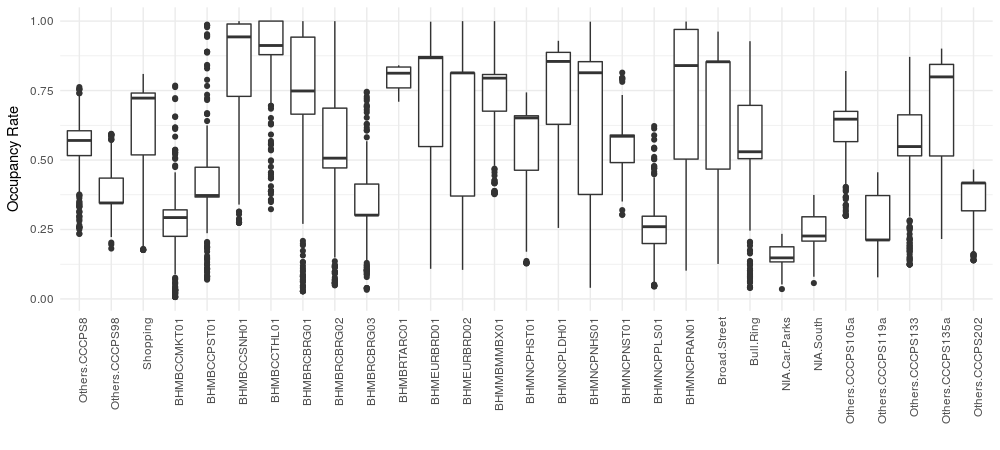}
  \caption{Occupancy rates summary of the 29 car parks operated by NCP in Birmingham, UK}
  \label{figure:occupancy-rates}
\end{figure}

To continue with the validation of our proposal in a real problem, we study the MAE prediction using the occupancy rates of the 29 car parks operated by NCP (National Car Parks) in Birmingham, UK, presented by Stolfi et al.~\cite{stolfi2017predicting}. Figure~\ref{figure:occupancy-rates} summarizes the occupancy rates of each car park.

We sampled the hyper-parameter space defined in Table~\ref{table:parameters}. In this case, we set the activation function of the output layer to be a $sigmoid$, because the occupancy rates range from 0 to 1. Therefore, the $MAE \in [0,1]$. Then, we fitted a truncated normal distribution in the range [0,1] to the samples and we calculated $p_{0.01}$. 

We sorted the architectures into deciles using the probability computed, trained ten architectures from each decile using the Adam optimizer
and computed the MAE. Table~\ref{table:birm-bp-corr} presents the correlation between the trained and MAE random sampling results. Again, there is a moderate to strong negative correlation~\cite{rubin2012statistics} between the approximation parameters (mean, sd, and log $p_{0.01}$) and the \emph{training} results. Particularly, the correlation between the training results and the calculated $p_{0.01}$ is negative. Therefore, the results are aligned with the previous ones, i.e., a good training result is correlated to a high probability.

\begin{table}[!h]	
    \caption{Correlation between the MAE of an RNN trained and the random sampling (car parks)}    
    \label{table:birm-bp-corr}
    \centering    
    \begin{tabular}{ lrrrrr }
    \toprule
    Epochs & NC & LB & Mean & Sd & log $p_{0.01}$ \\
    \midrule
    100 & 0.585 & -0.034 & 0.619 & -0.214 & -0.555 \\    
    \bottomrule
    \end{tabular}
\end{table}

\begin{figure}[!h]
  \centering
  \includegraphics[width=\textwidth]{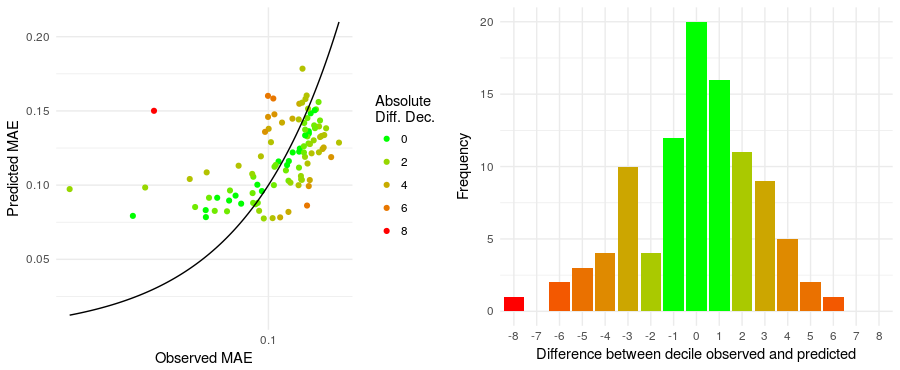}
  \caption{Difference between the predicted and the observed MAE and decile for the occupancy rates}
  \label{figure:birm-pred-obs}
\end{figure}

Using a uniform sample of the architectures (80\%), we fitted a linear model to the MAE based on $p_{0.01}$. Then, we predicted the MAE of the rest of the architectures (20\%) and compared the observed and predicted deciles. We repeated the process 30 times. On average, the residual standard error was equal to 0.034, while the Spearman correlation coefficient between the observed and the predicted deciles was equal to 0.584 ($p$-value=0.039). Also, the results showed that 50\% of the predicted MAE values have a difference of at most one decile to the observed value. 
Figure~\ref{figure:birm-pred-obs} presents the summarized results of one model. Note that the table includes the predicted values, as well as the ones used to fit the model.

The results show that the MAE sampling may be useful to predict the \emph{performance} of an RNN architecture, but more important than that in our opinion, the results support our claim that by performing a \emph{simple} MAE random sampling works well to rank a set of RNN architectures.

\subsection{Appliances Energy Consumption}

\begin{figure}[!h]
  \centering
  \includegraphics[width=\textwidth]{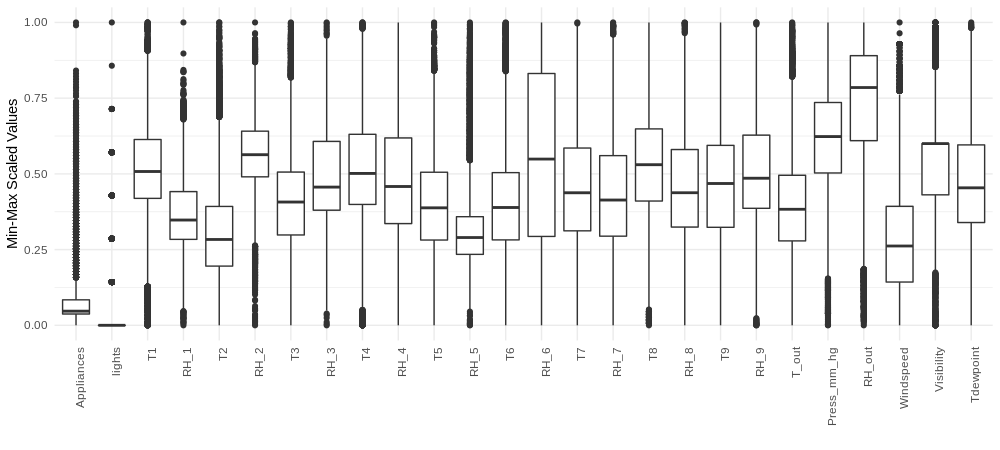}
  \caption{Appliances energy consumption and nearby weather data MinMax-scaled in the range [0,1]}
  \label{figure:energy-scaled}
\end{figure}

Finally, we tested our proposal using another real problem from a different domain, the appliances energy consumption of a house in Stambruges, Belgium, and the nearby weather data. This dataset was introduced by Candanedo et al.~\cite{candanedo2017data}. Figure~\ref{figure:energy-scaled} summarizes the energy consumption and the weather data scaled (MinMax) in the range [0,1]. Note that we use all the variables (26) as input to predict the \emph{appliances} and \emph{lights} consumption.

We sampled the hyper-parameter space (Table~\ref{table:parameters}), and we set the activation function to be a \emph{sigmoid}. Therefore, we approximated the MAE random sampling using a truncated normal distribution in [0,1], and we calculated $p_{0.01}$.

Then, we repeated the process, i.e., we sorted the architectures into deciles, trained ten architectures from each decile, and computed the MAE. Table~\ref{table:energy-bp-corr} presents the correlation between the trained and MAE random sampling results. Again, in this case, there is a moderate to strong negative correlation~\cite{rubin2012statistics} between the approximation parameters (mean, sd, and log $p_{0.01}$) and the \emph{training} results. 

\begin{table}[!h]	
    \caption{Correlation between the MAE of an RNN trained and the random sampling (appliances energy consumption)}    
    \label{table:energy-bp-corr}
    \centering    
    \begin{tabular}{ lrrrrr }
    \toprule
    Epochs & NC & LB & Mean & Sd & log $p_{0.01}$ \\
    \midrule
    100 & -0.246 & 0.103 & -0.278 & -0.502 & -0.383 \\    
    \bottomrule
    \end{tabular}
\end{table}

\begin{figure}[!h]
  \centering
  \includegraphics[width=\textwidth]{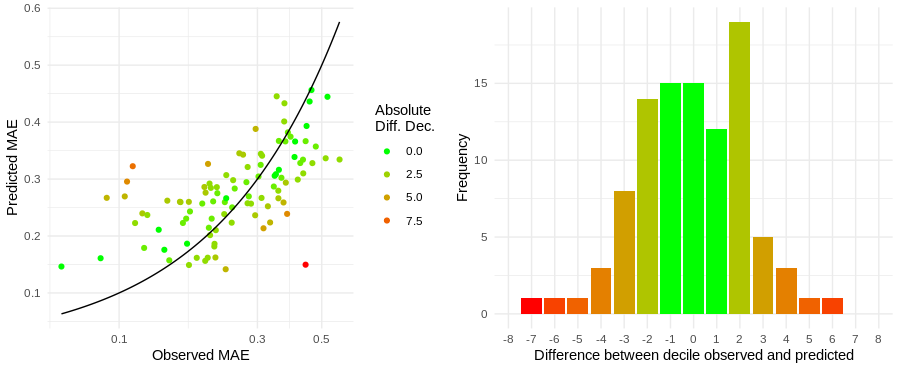}
  \caption{Difference between the predicted and the observed MAE and decile for the appliances energy consumption problem}
  \label{figure:energy-pred-obs}
\end{figure}

To conclude our experimentation, we uniformly sampled the architectures (80\%), and we fitted a linear model to the MAE based on $p_{0.01}$. Then, we predicted the MAE of the rest of the architectures (20\%) and compared the observed and predicted deciles. We repeated the process 30 times. On average, the residual standard error was equal to 0.042, while the Spearman correlation coefficient between the observed and the predicted deciles was equal to 0.500 ($p$-value=0.051). The results also showed that 43\% of the predicted MAE values have a difference of at most one decile to the observed value. 
Figure~\ref{figure:energy-pred-obs} presents the summarized results of one model. Note that the summary includes the predicted values, as well as the ones used to fit the model.

\section{Threats to Validity}\label{section:discussion}

In this section, we discuss the main advantages and drawbacks of our proposal, aiming to motivate the community to explore and expand the \emph{random sampling} as an alternative for hyper-parameter optimization. 

\begin{figure}[!h]
  \centering
  \includegraphics[width=0.6\textwidth]{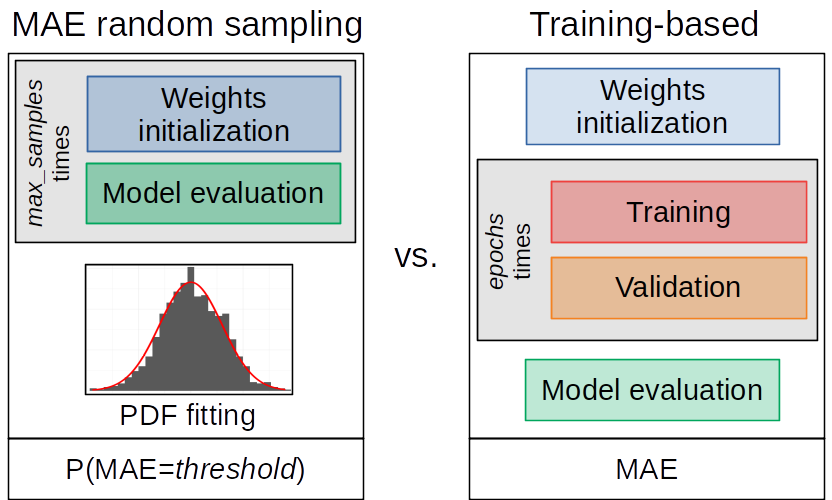}
  \caption{MAE random sampling versus training-based architecture performance evaluation}
  \label{figure:mrs-vs-training}
\end{figure}

Without a doubt, the most noticeable advantage of our proposal is the fact that no learning/training is needed to evaluate the expected performance of a given architecture. Thus, the computational resources needed to perform a hyper-parameter optimization might be kept low and the time might decrease dramatically, enabling the exploration of wider parameter space. Figure~\ref{figure:mrs-vs-training} outlines a high-level comparison of the MAE random sampling and a training-based architecture performance evaluation. Since training is costly, there have been proposed methods that limit the training effort~\cite{Domhan2015} or parallelize and distribute the execution. However, the training itself, regardless of the number of epochs (i.e., the number of times the process is repeated), requires complex computations (i.e., time and memory).

Another remarkable benefit (properly, a side effect) is that the \emph{best weights} found during the random sampling can be used as a start point to perform a training process (gradient-descent or evolutionary-based), improving the result.

However, in spite of its apparent simplicity, why the random sampling works, remains unexplained. The evidence presented in this study acts in favor of the usefulness of this method to predict the performance of an RNN (with its limitations), but this approach lacks a mathematical explanation (note that also ANN lacks it). Furthermore, the results presented are empirical and constrained to a regression problem (3 cases). Therefore, we can not state that our approach is valid for other problems, not even for other architecture types. But, to be fair, despite these important drawbacks, the evidence still supports our claim.


Then, is this approach worthy to give it a try? We really believe that the answer is yes! The random sampling is simple, yet powerful, and it presents a paradigm change for hyper-parameter optimization. But, to succeed, we envision that further study has to be made, from exploring different PDF fitting to alternative weight initialization or including different architecture types.

\section{Conclusions and Future Work}\label{section:conclusions}

In this study, we introduce a technique to rank stacked RNN architectures based on the probability of finding a set of weights whose MAE (given an input time series) is below a defined threshold. To estimate this probability we fit a truncated normal distribution to the results of a random sampling of the MAE, i.e., given a specific architecture we compute the MAE multiple times using randomly initialized weights with a normal distribution, and then, we fit the referred PDF to the results.

We test our proposal using a set of one hidden layer stacked RNN architectures (up to 100 LSTM cells in the hidden layer), and three problems: a sine wave, the occupancy rate of 29 car parks in a city, and the appliances energy consumption of a house. The results show that there is a moderate to strong negative correlation between the estimated probability of finding a set of weights whose MAE is less than or equal to a defined threshold and the MAE \emph{observed} after training a network using Adam optimizer. In other words, as the estimated probability increases, the MAE observed after training a network is likely to decrease.

By fitting a linear model based on the estimated probabilities to the \emph{observed} errors, independently for each problem, we show that the notions presented in this study may also be used to predict the MAE (after training the network with a gradient-descent method) of a sampled architecture. Thus, the MAE random sampling can be used as a heuristic to optimize the architecture of a network.

We discuss the main advantages and drawbacks of our proposal, concluding that our approach presents an interesting and promising new research line. Moreover, we introduce the research directions that we envision to be a priority.

Overall, the results indicate that it is possible to use the proposed approach as a \emph{low-cost}, \emph{training-free}, rule of thumb to compare the expected performance of RNN architectures.

As future work, we propose to test our proposal using deeper stacked RNN architectures, as well as other types of networks. We also propose to study the relation between the estimated probability and the number of neurons, with a special interest in finding if there is a point where the number of neurons \emph{saturates} the probability $p_t$ (plateaus, maximums, and minimums). Finally, we will study the application of our approach to actually optimizing a DNN.

\subsubsection*{Acknowledgments}

This research was partially funded by Ministerio de Econom\'ia, Industria y Competitividad, Gobierno de Espa\~na, and European Regional Development Fund grant numbers TIN2016-81766-REDT (\url{http://cirti.es}), and TIN2017-88213-R (\url{http://6city.lcc.uma.es}). European Union’s Horizon 2020 research and innovation programme under the Marie Skłodowska-Curie grant agreement No 799078. 
Universidad de M\'alaga, Campus Internacional de Excelencia Andaluc\'ia TECH.

\bibliographystyle{elsarticle-harv} 
\bibliography{low_cost}
\end{document}